\documentclass[12pt]{article}

\usepackage{scicite}
\usepackage{url}
\usepackage{interval}
\usepackage{threeparttable}
\usepackage{hyperref}
\usepackage{enumerate}
\usepackage{comment}
\usepackage{amsfonts}
\usepackage{amsmath}
\usepackage{mathtools}
\usepackage{nccmath}
\usepackage{times}

\newcommand{\N}{\mathbb{N}}

\newcommand{\R}{\mathbb{R}}

\usepackage[utf8]{inputenc}
\usepackage[english]{babel}
\usepackage[T1]{fontenc} 
\usepackage{fouriernc} 
\usepackage{etoolbox}
\usepackage{bm}
\patchcmd{\section}{\bfseries}{\bfseries\boldmath }{}{}
\patchcmd{\subsection}{\bfseries}{\bfseries\boldmath }{}{}

\title{Performance Prediction in Major League Baseball by Long Short-Term Memory Networks}

\author
{Hsuan-Cheng Sun,$^{1\ast}$ Tse-Yu Lin,$^{2}$ Yen-Lung Tsai$^{3}$\\
\\
\normalsize{$^{1}$Center for Information Systems and Technology,}\\
\normalsize{Claremont Graduate University, Claremont, USA}\\
\normalsize{$^{2}$Data Science Degree Program, National Taiwan University, Taipei, Taiwan}\\
\normalsize{$^{3}$Department of Mathematical Sciences, National Chengchi University, Taipei, Taiwan}\\
\\
\normalsize{$^\ast$To whom correspondence should be addressed; E-mail:  hsuan-cheng.sun@cgu.edu.}
}

\date{}

\begin{document} 


\baselineskip18pt


\maketitle


\begin{abstract}
 Player performance prediction is a serious problem in every sport since it brings valuable future information for managers to make important decisions. In baseball industries, there already existed variable prediction systems and many types of researches that attempt to provide accurate predictions and help domain users. However, it is a lack of studies about the predicting method or systems based on deep learning. Deep learning models had proven to be the greatest solutions in different fields nowadays, so we believe they could be tried and applied to the prediction problem in baseball. Hence, the predicting abilities of deep learning models are set to be our research problem in this paper. As a beginning, we select numbers of home runs as the target because it is one of the most critical indexes to understand the power and the talent of baseball hitters. Moreover, we use the sequential model Long Short-Term Memory as our main method to solve the home run prediction problem in Major League Baseball. We compare models' ability with several machine learning models and a widely used baseball projection system, sZymborski Projection System. Our results show that Long Short-Term Memory has better performance than others and has the ability to make more exact predictions. We conclude that Long Short-Term Memory is a feasible way for performance prediction problems in baseball and could bring valuable information to fit users' needs.
\end{abstract}


\section{Introduction}
\label{intro}

Nowadays, data analytics become more critical since teams in professional sports usually spend multi-million dollars on a single decision \cite{Schumaker2010}, so the data mining applications can help them and lower the risks. The predictions and results from data analytics could bring advantages to the domain users and fit their needs. Especially, player performance prediction is one of the most important problems in the sports domain. Those predictions can provide additional information for managers to make correct decisions. For example, they may need the information to find the potential of players to sign contracts, issue trades with other teams, and set different training programs for players. Therefore, a good prediction method can bring lots of benefits and make a team being powerful. 

In baseball, after Bill James popularized the term "sabermetrics", people started to put more emphasis on analyzing baseball statistics \cite{baumer2014sabermetric}. Analysts are dedicated to finding useful information in baseball data and build prediction systems, which are also called projection systems, to get the more accurate future performance of players. There are some common ways to make predictions, such as comparing players with similar players' historical performance \cite{silver2003introducing} and weighting differently for past data \cite{ttangoMarcel}. In recent years, since more analytical tools are developed and advanced data can easily get, projection systems are able to add more information in and make more accurate predictions \cite{dszymborskiZips}. Furthermore, machine learning skills are the popular models to be used to build the supported system. This trend appeared in different sports field, such as in football \cite{al2018decision} \cite{herold2019machine}, and there were several researches based on machine learning in baseball predictions \cite{hamilton2014applying} \cite{karnuta2020machine}.

Furthermore, nowadays, deep learning, which is also called the deep neural network, is one of the most robust methods in Artificial intelligence (AI) used in a variety of fields, such as object detection \cite{Redmon_2016}, natural language processing  \cite{Peters_2018} and image classification \cite{He_2016}.  By good devices like graphic processing unit (GPU) and larger datasets, deep learning can be trained as a great function to solve problems. For sequential data tasks, Recurrent
Neural Networks (RNN, \cite{Rumelhart:1986we}) and Long Short-Term Memory (LSTM) \cite{10.1162/neco.1997.9.8.1735} have proven to be very useful methods, such as knowing tomorrow's weather by information of past $n$ days \cite{qing2018hourly}, output a sentence from the previous one \cite{sutskever2014sequence}, and generating a paragraph from a topic input \cite{radford2019language}. Their structure can handle information from every past time input and has a strong ability to obtain the knowledge behind the sequence. 

Therefore, we are curious about how deep learning can be used to predict the future performance of baseball players by inputting their past performance. However, in the literature review of machine learning applications in baseball (2017), Koseler and Stephan mentioned that there is only 3 out of 32 cases of baseball analysis research included artificial neural networks as their method \cite{koseler2017machine}. In the 3 cases, only Lyle (2007) focused on players' performance \cite{lyle2007baseball}. He used artificial neural networks (ANN) to predict six hitters' stats and compare them with existing projection systems. Lyle's prediction can only show that prediction by ANN on triple hits outperformed other methods. Due to lack of analysis, we would like to have more researches on performance prediction via deep learning and make it a stable projection system in the future. Furthermore, we want to provide more valuable information from the study to help domain users. The actionable knowledge we discover could be support for the problem-solving materials for these professionals. Our research aims to help the users in MLB industries and bring more benefits for them.

As a beginning study for the topic, we select numbers of home runs (HR) to make predictions since it is one of the most critical indexes to understand the power and the talent of baseball hitters. In this paper, we focus on predicting individual baseball players' home run performance in Major League Baseball (MLB) by using models based on the Long Short-Term Memory structure. To evaluate their capacity, we analyze and compare the results of our models with an accurate projection system, called sZymborski Projection System (ZiPS) \cite{dszymborskiZips} and other traditional machine learning methods. Hence, in this paper, we have the following contributions:
\begin{itemize}
	\item We started a new research direction on using deep learning to predict baseball players' future performance.
	\item We have a systematic analysis of predictions to build new insight into the problem.  
	\item Our result demonstrated that deep learning could be a better solution to solve the performance prediction problem.
	\item With our study and results, we believe the information would be helpful for the domain users and could create actionable knowledge as the support for them. 
\end{itemize}

In the rest of the paper, section 2 introduces the background knowledge of related works, baseball projection system, and LSTM, section 3 shows details about our experiments, such as dataset and research problem, section 4 contain our analysis on the results and discussion about our paper, and section 5 is for the conclusion.

\section{Background}

In this section, we show previous researches and explain some widely used projection systems in the baseball industry. Since a good projection system could be really complicated and has huge business benefits, almost all of the authors do not express how the system works clearly, but we would try to explain to them as detailed as possible. Also, we would explain our main method, LSTM, in this section.

\subsection{Related Works}
For the literature review, we present previous studies on several relative topics. In previous research on baseball performance prediction, Brown (2008) studied players’ batting average during a single season \cite{brown2008season}. He used the first-half season (3 months) batting average data of players to predict their performance in the second half. He confirmed that performance in the first-half season is useful information for predicting their second-half performance. It is useful because it brings managers a clear view of players’ in-season performance. Moreover, Jiang, and Zhang (2010) further boosted Brown’s research \cite{jiang2010empirical}. They focused on different kinds of empirical Bayes methods and predicted in-season players’ performance with those methods like Brown did. With a comprehensive comparison of these methods, they found empirical Bayes methods were better than least-squares predictor. Their discoveries provided later researchers with more great methods. Lyle (2007) had built a new prediction model for MLB players based on ensemble learning skills and used players’ past accumulated stats in 162 games (1 year) to predict their future performance. After comparing results on six offensive indies with other machine learning ways and existing prediction systems, he observed that none of the individual systems outperformed the others in all of the six stats. Finally, he concluded each item has its own suitable prediction method. Researchers have to do more study on this topic to get clearer answers. 

On the other hand, for home run predictions, some researchers focus on the other information of players instead of focusing only on the stats as our paper. Sawicki, Hubbard, and Stronge (2003) put emphasis on the ball and bat \cite{sawicki2003hit}. They studied the physical situations that the ball meets the bat, and they consider every factor that would influence the ball and calculated how far it could travel. Their results were useful to modify and optimize the flight models. In their conclusion, they also found an optimal curveball could go further than other kinds of balls. Goldschmied, Harris, Vira, and Kowalczyk (2014) paid attention to the biological information of players \cite{goldschmied2014drive}. They examined the performance of historical players who faced career milestones, the 500 or 600 HR, and found that their performance would be poor under this kind of pressure and matched the biological theory. Moreover, their result showed the players who took performance-enhancing medicine might not be affected by this biological influence. 

\subsection{Projection Systems in MLB}


Marcel the Monkey Forecasting System (Marcel) is a projection system published in 2013 \cite{ttangoMarcel}. It is considered to be the father of the projection system in MLB. Although it was designed in a simple way, its predicting results are still reliable. The author firstly used past three seasons’ data with heavier weight for the most recent season to count base numbers. Then the number would be adjusted by league average stats. Finally, the prediction would be made after considering age factors.

Player Empirical Comparison and Optimization Test Algorithm (PECOTA) is another projection system proposed by Nate Silver and is owned by Baseball Prospectus now \cite{silver2003introducing}. The system also counts a baseline with the player's past data with heavier weight for the most recent season. In the next step, PECOTA will find similar historical players with new players' baseline and other body information. Historical stats would be weighted differently and create the prediction.

Steamer \cite{jcrossSteamer} and ZiPS are projection systems shown on FanGraphs, one of the largest websites for baseball analysts around the world. Steamer is created by Jared Cross, Dash Davidson ,and Peter Rosenbloom. Similar to Marcel, Steamer predicts future performance based on the historical data with different weights. For Steamer, weights are produced by regression analysis instead of fixed weights in Marcel.

ZiPS is the abbreviation of sZymborski Projection System and was developed by Dan Saymborski. Similar to PECOTA, ZiPS create a baseline by weighting heavier for more recent seasons. It takes four years of data for normal players and three for those very young and very old. The baseline and other information, like velocity and pitching data, are used to find similar historical players to predict player performance. In this paper, we compare our result to ZiPS since it is considered to be one of the most accurate projection systems and its prediction data could be accessed easily.

\subsection{Long Short-Term Memory}

Recurrent-type neural networks (RNN) are widely applied to address practical problems involving sequential data \cite{raza2021news}. Recurrent-type neural networks (RNN) is one of the classes of neural network models that can capture hidden information in the sequential data by its recursive structure. Long Short-Term Memory (LSTM) is based on RNN and designed to solve unstable training progress in RNN. It also enhances the performance of RNN by the ``gate'' and memory state structure. Nowadays the projection systems use players' past performance to make their future predictions, so we see the problem as a sequential type and would like to apply it to LSTM.

Here we give a brief explanation about how Vallina RNN, which is the original RNN, works.

RNN is one or more layers that contain several RNN cells each. Let $x=\{x_t\}_{t=1}^T$ be a sequential data in $\R^n$, and $m\in\N$ be the dimension of RNN layer. The layer is a function $\varphi:\R^m\times\R^n\to\R^m$ which output the hidden information $h_{t}$ at time $t$ defined recursively as follows:
\begin{ceqn}
	\begin{align*}
	h_{t} & = \varphi(h_{t-1}, x_{t})\\
	& = \tanh(W_h\cdot h_{t-1}+W_x\cdot x_{t}+b)
	\end{align*}
\end{ceqn}

where $h_{t-1}$ is the vector of cell status at time $t-1$, $W_{h}\in\R^{m\times m}$ and $W_{x}\in\R^{n\times n}$ are tunable parameters with respect to $h_{t-1}$ and $x_t$, and $b\in\R^m$ is the bias of the layer. $\tanh:\R \rightarrow (-1,1) ,~ 
{\rm tanh}(x)={\frac {(e^{x}-e^{-x})}{(e^{x}+e^{-x})}}$ is the hyperbolic tangent function which acts component-wisely. The final hidden state $h_T$ will be the input for the next layer or simply the output. 

However, the structure of RNN led to gradient vanishing \cite{Graves:1503877} which is the reason for overfitting in training progress. LSTM was announced to solve the problem. Similar to RNN, for each LSTM layer at time $t$ can be written as follow:
\begin{ceqn}
	\begin{align}
	\ f_{t} & = \sigma(W_{hf}\cdot h_{t-1}+W_{xf}\cdot{ x}_{t}+b_{f})\\
	\ i_{t} & = \sigma(W_{hi}\cdot h_{t-1}+W_{xi}\cdot{ x}_{t}+b_{i})\\
	\ o_{t}& = \sigma(W_{ho}\cdot h_{t-1}+W_{xo}\cdot{ x}_{t}+b_{o})\\
	{\tilde C}_{t} & = \tanh(W_{hc}\cdot{ h}_{t-1}+W_{xc}\cdot{ x}_{t}+b_{c})\\
	C_{t} & = \ f_{t}\cdot C_{t-1}+i_{t}\cdot\tilde{C_{t}}\\
	h_{t} & = \ o_{t}\cdot\tanh(C_{t})
	\end{align}
\end{ceqn}

where $f$, $i$, $o$, ${\tilde C}$ are {\it forget gate}, {\it input gate}, {\it output gate} and {\it cell memory} at time $t$, respectively. $C$ is the cell memory, $W_{h\star}\in\R^{m\times m}$ and $W_{x\star}\in\R^{n\times n}$ are the metric of trainable parameters and $b_{\star}\in\R^m$ is the bias, where $\star\in\{f, i, o, {\tilde C}\}$. $\sigma: \R \rightarrow (0,1) ,~ \sigma(x)={\frac {1}{1+e^{-x}}}$ is sigmoid function which acts component-wisely. Three gates serve as weights to help the model capture the information from the past and solve the unstable issue. 

\section{Experiments}

\subsection{Dataset Preparation}

We collect a total of 5401 players recorded in MLB from 1961 to 2019 as our dataset from the website Baseball-Reference.com \cite{BR}. It contains detailed information for MLB players. Most of the baseball activities are recorded on the website. For more information, readers can check the link: https://www.baseball-reference.com/. 

For each year, we take 21 features from the players. Table \ref{table:feature} shows all features used in this paper. Not only their performance in the field, but we also consider body information, such as age, height, and weight.
Besides, since there are too many players who hit zero homers, we exclude data if players have less than 50 plate appearances of the season with zero home runs. 

\begin{table}[h!]
	\centering
	\begin{tabular}[t]{|c|c|c|}
		\hline
		Feature & Feature & Feature \\\hline
		\hline
		Age & Runs Scored & Intended Bases on Balls\\\hline
		Height & Double Hits & Grouned Into Double Plays\\\hline
		Weight & Triple Hits & Hit By a Pitch \\\hline
		Season & Stolen Bases & Plate Appearences \\\hline
		Home Run & Caught Stealing & Runs Battle In \\\hline
		Hit & Games Played & Bases on Balls \\\hline
		StrikeOut & Sacrifice Flies & Sacrifice Hits\\\hline
	\end{tabular}
	\caption{Features used in this paper.}
	\label{table:feature}
\end{table}

\subsection{Problem Description}

\begin{figure} [htbp]
	\begin{center}
		\includegraphics[width=8cm]{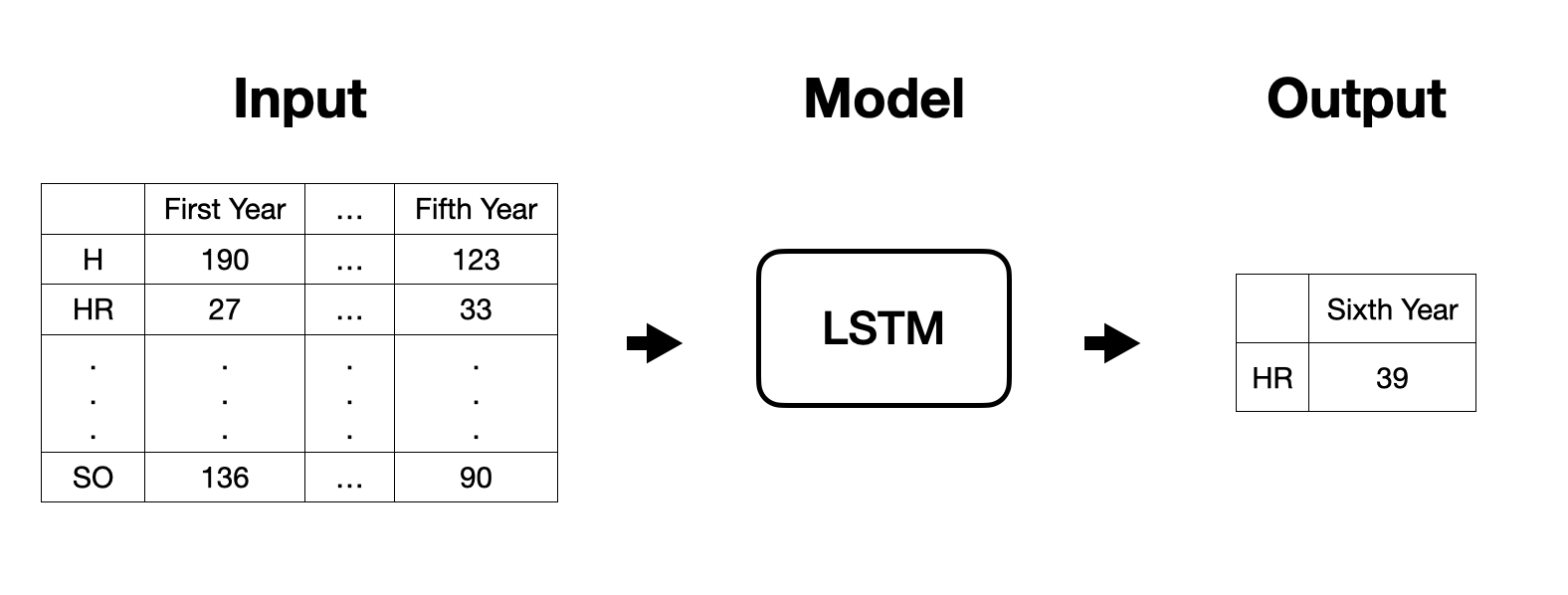} 
		\caption{Simple example of the problem in the paper} 
		\label{fig:inout} 
	\end{center}
\end{figure}

In this paper, we would like to use LSTM to predict home run numbers in the future. We will take 6 continuous years of data and input the first 5 continuous-years status with 21 features per year and predict the home run in the sixth year. Therefore, each data point is the form
$$
(\{x_t\}_{t=j}^{j+4}, y_{j+5}),
$$
where $x_j\in\mathbb{R}^{21}$ is the status at year $j$, and $y_{j+5}\in\{0,~1, \cdots,74\}$, is the HR of $x_{j+5}, 1961\leq j\leq 2014$. Figure \ref{fig:inout} shows a simple visual example.

We take data points from players' career data like a moving window. For example, if a player plays from 2011 to 2018, then we can take 3 data points, start from the first six years, then the second six years, and the last six years. We use data from 1961 to 2017 as training data and data points in 2018 and 2019 are the testing data. In other words, data points include $y_{2018}$ and $y_{2019}$ are our testing data, the rest of them are training. Since we find that all models' performance drop when using original training data to predict data points in 2019, we add data points in 2018 to the training data and retrain the model when predicting data points in 2019. There are 9828 data points from 1961 to 2017, 184 data points in 2018, and 191 data points in 2019. For ZiPS, it has 449 and 485 predictions in 2018 and 2019, respectively. We show the numbers in Table \ref{table:datapt}.
\begin{table}[h!]
	\centering
	\begin{tabular}[t]{|c|c|c|}
		\hline
		& 2018 & 2019 \\\hline
		\hline
		Training & 9828 & 10006\\\hline
		Testing & 184 & 191\\\hline
		ZiPS & 449 & 485\\\hline
	\end{tabular}
	\caption{Data points for training and testing}
	\label{table:datapt}
\end{table}

\subsection{Prediction Models}

In this paper, we use five kinds of LSTM models base on previous research \cite{sun2019LSTM}. Table \ref{table:LSTM_arch} shows our main architecture. 

\begin{table}[h!]
	\centering
	\begin{tabular}[t]{|c|cccccc|c|}
		\hline
		& LSTM & D/BN & TD & FC & D/BN & FC & param.\\
		\hline
		A & 128$*$128 & 0.5 & & 1024 & 0.5 &  & 865,793\\
		B & 32$*$16$*$8 & 0.5  & & 512 & 0.5 & 64& 64,737\\
		C & 32$*$32 & $\surd$ & & 512 & $\surd$ & 64 & 132,737\\
		D & 64$*$64 & 0.5 & $\surd$ & 512 & 0.5 & 64& 114,699\\
		E & 32$*$32 & 0.5 & & 512 & 0.5 & 64 & 130,561\\
		\hline
	\end{tabular}
	\caption{LSTM model architechture. BN: batch normalization; D: dropout; FC: fully connected layer; TD: timestep-wise dimension reduction; LSTM: long short-term memory layer; In LSTM column, $p*s$ means a layer with $p$ cells follows a the other one with $s$ cells. In D/BN columns, number represents dropout rate and $\surd$ means BN is used instead of dropout.}
	\label{table:LSTM_arch}
\end{table}

Basically, we use several LSTM layers in the beginning and follow a fully connected neural network. Each model contains either batch normalization \cite{ioffe2015batch} or dropout \cite{srivastava2014dropout} layers to stable training process. In model D, we use timestep-wise dimension reduction to reduce dimensions before standard fully-connected layers. Finally, we have one neuron with ReLU activation function \cite{conf/icml/NairH10} for the output. For these hyperparameters, we have tried 8, 16, 32, 64, 128, and 256 for LSTM cells and 1 to 3 layers in LSTM. We found that the cells should not be more than 128 and layers should not exceed 3, or models would be overfitting. For fully connected layers, we also try 1 to 3 layers and 64, 128, 256, 512, and 1024 for neurons in a layer. This time we found that two layers are enough and too many parameters would decrease the accuracy rate. Hence, the models in the Table \ref{table:LSTM_arch} are the best combinations in our experiments.

We also consider several state-of-the-art sequential models in this paper, such as gated recurrent unit (GRU) \cite{cho2014learning}, bidirectional LSTM (BiLSTM) \cite{graves2005framewise}, and attention-based LSTM (AT-LSTM) \cite{bahdanau2014neural}. We have created double GRU layers with 64 cells followed by fully connected layers that contain 512 and 64 neurons to see their performance. We use a similar structure to test BiLSTM by switching the double GRU layers to bidirectional LSTM layers with 64 cells. Finally, we put attention mechnism between a bidirectional LSTM layer and an LSTM layer with fully connected networks to evaluate the performance of AT-LSTM. However, we do not test their different structure due to time limitations. We would put them into the limitation and leave the part as future research directions. Furthermore, we did not include other novel sequential models, such as transformer \cite{vaswani2017attention} and convolution-neural-network-based models \cite{li2018news}, in the paper since we are still studying them. They are good research directions on the topic.

We train all models for 1000 epochs with a learning rate $10^{-3}$. Mean square error is set as our loss function and we set adam \cite{kingma2014adam} be the optimizer. In this paper, we also compare model performance with linear regression \cite{seber2012linear}, Support Vector Machines \cite{10.1109/5254.708428}, Random Forests \cite{RF}, and a simple fully connected neural network \cite{Jain96artificialneural}. Later, we simply show linear regression as LR, Support Vector Machines as SVM, Random Forests as RF, and a fully connected neural network as NN. For RF, we use 200 estimators. For NN, we construct a three-layer model with 1024, 512, 128 neurons, respectively, and train for 1000 epochs. 

\section{Results and Discussions}
\subsection{Loss and Accuracy Rate}

\begin{table}[h!]
	\centering
	\begin{tabular}[t]{|ccccc|}
		\hline
		& 2018 & 2018 & 2019 & 2019  \\
		Models & MAE & RMSE & MAE & RMSE \\
		\hline
		ZiPS & 6.704 & 8.27 & 6.737 & 9.381 \\
		\hline
		LR & 5.205 & ${\bf6.587}$  & 6.703 & 8.931 \\
		SVM & 6.707 & 8.854 & 7.78 & 10.753 \\
		RF & 7.332 & 9.404 & 8.398 & 11.067 \\
		NN & 13.293 & 16.537 & 15.178 & 18.982 \\
		\hline
		LSTM A & 5.519 & 7.424 & 6.721 & 8.824 \\
		LSTM B & 5.783 & 7.724 & 6.393 & 8.57 \\
		LSTM C & 5.313 & 6.905 & ${\bf6.138}$ & ${\bf8.269}$ \\
		LSTM D & 5.202 & 6.895 & 6.973 & 9.497 \\
		LSTM E & ${\bf5.176}$ & 6.908 & 6.571 & 8.909 \\
		\hline
		GRU & 5.721 & 7.43 & 6.551 & 8.83 \\
		BiLSTM & 5.632 & 7.534 & 6.835 & 9.042 \\
		AT-LSTM & 5.576 & 6.973 & 6.584 & 8.691 \\
		\hline
	\end{tabular}
	\caption{MAE and RMSE of each methods in 2018 and 2019}
	\label{table:LOSS}
\end{table}

Table \ref{table:LOSS} shows our result on mean absolute error (MAE) and root-mean-square error (RMSE), which can be written as following:

\begin{ceqn}
	\begin{align}
	{\rm RMSE}(\theta)&=\sqrt{{\frac {\sum\limits _{i=1}^{k} ( y_{i}-f_\theta(x_i))^{2}}{k}}}\\
	{\rm MAE}(\theta)&={\frac {1}{k}}\sum\limits _{i=1}^{k} \left| y_{i}-f_\theta(x_i)\right|
	\end{align}
\end{ceqn}
where $f_\theta$ denotes the model $f$ under the parameter set $\theta$, $x_i$ and $y_i$ are inputs and outputs, and $k$ is the total number of data points.

As we can see, LSTM models have great performance among all methods. Model E gets the lowest MAE in 2018 and model C has the lowest error rate in 2019. Other than LSTM models, linear regression and ZiPS perform better than other machine learning ways. Although linear regression gets the lowest RMSE in 2018, LSTM models still have a close rate to it. In 2018, LSTM models have close performance with linear regression, but in 2019, four of them outperform all the ways and the left one is close to linear regression, which shows their excellent predicting ability. Although GRU, BiLSTM, and AT-LSTM did not perform as well as LSTM models, they still have better performance than machine learning ways. We see they have great potentials for the predicting problem. Generally, LSTM models perform better than machine learning ways and ZiPS. 

For LSTM models, we can observe that fewer LSTM cell in a layer, like model C, performs better than those with more LSTM cells, like model A and D, so the model does not need to be too complicated. Model C and E show that dropout and batch normalization are both regularization methods in our task, and timestep-wise dimension reduction is also useful in our model. Model A, C, E are stable, while model D and B focus too much in 2018 and 2019, respectively.

\begin{table}[h!]
	\centering
	\begin{tabular}[t]{|cccccc|}
		\hline
		& 0 & $\interval{-1}{1}$ & $\interval{-3}{3}$ & $\interval{-5}{5}$ & $\interval{-10}{10}$  \\
		\hline
		ZiPS & 3.12 & 9.8  & 28.95 & 48.11 & 81.74 \\
		\hline
		Regression & 4.35 & 17.39 & 41.30 & 60.33 & ${\bf90.22}$ \\
		SVM & 8.7 & 14.67 & 36.41 & 52.17 & 78.80 \\
		RF & 5.43 & 17.39 & 32.07 & 45.65 & 74.46 \\
		NN & 2.72 & 9.78 & 15.22 & 23.91 & 44.57 \\
		\hline
		LSTM A & 11.96 & 21.74 & 45.11 & 58.15 & 85.33 \\
		LSTM B & ${\bf13.59}$ & 23.91 & 41.30 & 56.52 & 81.52 \\
		LSTM C & 10.87 & 21.74 & 40.76 & 58.70 & 89.13 \\
		LSTM D & 10.33 & 22.83 & ${\bf46.74}$ & ${\bf60.97}$ & 87.5 \\
		LSTM E & 12.5 & ${\bf24.46}$ & ${\bf46.74}$ & 60.87 & 87.5 \\
		\hline
		GRU & 10.33 & 20.65 & 38.59 & 58.15 & 83.7 \\
		BiLSTM & 11.41 & 23.37 & 42.93 & 58.15 & 82.07 \\
		AT-LSTM & 9.78 & 17.39 & 37.5 & 52.72 & 86.96 \\
		\hline
	\end{tabular}
	\caption{Accuracy rate in each difference interval in 2018}
	\label{table:results2018}
\end{table}

\begin{table}[h!]
	\centering
	\begin{tabular}[t]{|cccccc|}
		\hline
		& 0 & $\interval{-1}{1}$ & $\interval{-3}{3}$ & $\interval{-5}{5}$ & $\interval{-10}{10}$  \\
		\hline
		ZiPS & 5.15 & 14.43  & 36.91 & 51.75 & 77.94 \\
		\hline
		Regression & 3.14 & 14.66 & 38.74 & 51.31 & ${\bf80.1}$ \\
		SVM & 7.85 & 19.37 & 32.46 & 49.74 & 73.30 \\
		RF & 6.81 & 17.8 & 29.84 & 43.46 & 63.87 \\
		NN & 2.09 & 7.33 & 18.85 & 23.56 & 39.27 \\
		\hline
		LSTM A & 9.42 & 16.75 & 35.60 & 51.83 & 75.92 \\
		LSTM B & ${\bf12.57}$ & ${\bf30.94}$ & 38.22 & 53.93 & 76.96 \\
		LSTM C & 9.95 & 21.47 & ${\bf40.31}$ & 56.02& 79.58 \\
		LSTM D & 10.99 & 21.47 & 39.79 & 50.79 & 74.35 \\
		LSTM E & 11.52 & 21.47 & 38.22 & 51.83 & ${\bf80.1}$ \\
		\hline
		GRU & 10.99 & 16.75 & 37.7 & ${\bf58.12}$ & 78.53 \\
		BiLSTM & 9.95 & 14.66 & 36.65 & 51.31 & 75.92 \\
		AT-LSTM & 8.9 & 16.23 & 32.98 & 51.31 & 78.01 \\
		\hline
	\end{tabular}
	\caption{Accuracy rate in each difference interval in 2019}
	\label{table:results2019}
\end{table}

Next, we see the accuracy rate in each difference interval.  The difference means the difference between true home run numbers and predictions. If the difference locates in the interval, we see the prediction as to the correct prediction under this interval. For example, if a player hits 20 home runs and the prediction is 23, then the prediction is correct under the interval $\interval{-3}{3}$ but wrong under the interval $\interval{-1}{1}$. GT represents ground truth.

From Table \ref{table:results2018} and Table \ref{table:results2019}, 
LSTM models all reach about 10\% or more accuracy rates for predicting exactly the true home run numbers. About 20\% testing data could be predicted under the difference interval of 1, and LSTM B in 2019 even reaches a marvelous 30\% accuracy rate. All other machine learning ways and ZiPS are not able to get such a high accuracy rate in a small difference. Linear regression and ZiPS have similar performance with LSTM models when the difference is more than 3, but SVM, random forest, and neural network still perform poorly even the interval is huge. The observations above explain why LSTM models have lower MAE and RMSE than regression and ZiPS. Accuracy rates of GRU, BiLSTM, and AT-LSTM are between machine learning ways and LSTM models. Their performance may be enhanced by more studies and tests. Overall, ZiPS and traditional machine learning models can make correct predictions under a larger error range, but they can not predict exactly or in a small difference of the home run in the future. On the other hand, ZiPS not only shows its excellent performance in the small difference, but also includes more correct predictions under large differences than those machine learning models and ZiPS do. For the latter part, we decide to focus on LSTM models, machine learning ways, and ZiPS to emphasize our main problems. 

\subsection{Prediction Results by Class}

\begin{table}[h!]
	\centering
	\begin{tabular}[t]{|cccccc|}
		\hline
		& $\interval{0}{9}$ & $\interval{10}{19}$ & $\interval{20}{29}$ & $\interval{30}{39}$ & 40+ \\
		\hline
		LR & 14 & 11 & ${\bf5}$ & 1 & 0 \\
		SVM & 13 & 8 & 2 & ${\bf4}$ & 0 \\
		RF & 22 & 9 & 1 & 0 & 0 \\
		NN & 18 & 0 & 0 & 0 & 0 \\
		\hline
		LSTM A & 22 & 17 & 1 & 0 & 0 \\
		LSTM B & 20 & ${\bf22}$ & 2 & 0 & 0 \\
		LSTM C & ${\bf24}$ & 16 & 2 & 0 & 0 \\
		LSTM D & ${\bf24}$ & 16 & 2 & 0 & 0 \\
		LSTM E & 21 & 19 & ${\bf5}$ & 0 & 0 \\
		\hline
		GT & 75 & 62 & 34 & 11 & 2 \\
		\hline\hline 
		ZiPS & 20 & 15 & 5 & 4 & 0 \\
		\hline
		ZiPS GT & 273 & 97 & 56 & 21 & 2\\
		\hline
	\end{tabular}
	\caption{Number of correct predictions in 2018 with difference in 1.}
	\label{table:2018D1}
\end{table}

\begin{table}[h!]
	\centering
	\begin{tabular}[t]{|cccccc|}
		\hline
		& $\interval{0}{9}$ & $\interval{10}{19}$ & $\interval{20}{29}$ & $\interval{30}{39}$ & 40+ \\
		\hline
		LR & 15 & 12 & 0 & 1 & 0 \\
		SVM & 21 & 10 & 5 & 1 & 0 \\
		RF & ${\bf27}$ & 5 & 1 & 1 & 0 \\
		NN & 14 & 0 & 0 & 0 & 0 \\
		\hline
		LSTM A & 17 & 10 & 2 & ${\bf3}$ & 0 \\
		LSTM B & 12 & ${\bf23}$ & 4 & 1 & 0 \\
		LSTM C & 17 & 18 & ${\bf6}$ & 1 & 0 \\
		LSTM D & 26 & 8 & 4 & ${\bf3}$ & 0 \\
		LSTM E & 20 & 18 & 3 & 0 & 0 \\
		\hline
		GT & 70 & 60 & 35 & 20 & 6 \\
		\hline\hline
		ZiPS & 38 & 21 & 10 & 1 & 0 \\
		\hline
		ZiPS GT & 260 & 114 & 60 & 43 & 8 \\
		\hline
	\end{tabular}
	\caption{Number of correct predictions in 2019 with difference in 1.}
	\label{table:2019D1}
\end{table}

We list the prediction results in 2018 and 2019 under each difference interval. To be checked easily, we simply spilt home runs into five classes. Since ZiPS make more predictions than our dataset, we list its result separately.

Table \ref{table:2018D1} and \ref{table:2019D1} show the result in 2018 and 2019 under difference 1. We can find that the most correct predictions concentrate is under 19 home runs. Correct predictions in 20-29 home runs drop and LSTM models perform better than other ways. In 30-39 home runs, LSTM models could make 1 or 3 correct predictions. Surprisingly, SVM in 2018 could make 4 correct ones. No method can have correct predictions on the players who can hit more than 40 home runs. From these two tables, we can find LSTM models performed well generally in such small difference. It is a challenge for machine learning models, but sometimes their result would be better than LSTM. Although ZiPS have more analyzed samples, its accuracy is low in two years. In addition, the results show players who hit under 19 HRs are more predictable than those who hit more, and predicting performance for players who can hit at least 20 HRs is still a problem.

\begin{table}[h!]
	\centering
	\begin{tabular}[t]{|cccccc|}
		\hline
		& $\interval{0}{9}$ & $\interval{10}{19}$ & $\interval{20}{29}$ & $\interval{30}{39}$ & 40+ \\
		\hline
		Regression & 36 & 29 & ${\bf9}$ & 2 & 0 \\
		SVM & 30 & 23 & 8 & ${\bf5}$ & ${\bf1}$ \\
		RF & 41 & 16 & 2 & 0 & 0 \\
		NN & 38 & 0 & 0 & 0 & 0 \\
		\hline
		LSTM A & 43 & ${\bf36}$ & 3 & 0 & 0 \\
		LSTM B & 39 & 35 & 2 & 0 & 0 \\
		LSTM C & 43 & 26 & 6 & 0 & 0 \\
		LSTM D & ${\bf46}$ & 32 & 5 & 1 & 0 \\
		LSTM E & 42 & 35 & ${\bf9}$ & 0 & 0 \\
		\hline
		GT & 75 & 62 & 34 & 11 & 2 \\
		\hline\hline
		ZiPS & 67 & 35 & 21 & 7 & 0 \\
		\hline
		ZiPS GT & 273 & 97 & 56 & 21 & 2\\
		\hline
	\end{tabular}
	\caption{Number of correct predictions in 2018 with difference in 3.}
	\label{table:2018D3}
\end{table}

\begin{table}[h!]
	\centering
	\begin{tabular}[t]{|cccccc|}
		\hline
		& $\interval{0}{9}$ & $\interval{10}{19}$ & $\interval{20}{29}$ & $\interval{30}{39}$ & 40+ \\
		\hline
		Regression & 38 & ${\bf38}$ & 7 & 1 & 0 \\
		SVM & 31 & 15 & ${\bf13}$ & 2 & ${\bf1}$ \\
		RF & 39 & 11 & 4 & 2 & 0 \\
		NN & 36 & 0 & 0 & 0 & 0 \\
		\hline
		LSTM A & 36 & 18 & 10 & ${\bf4}$ & 0 \\
		LSTM B & 33 & 31 & 6 & 3 & 0 \\
		LSTM C & 35 & 28 & 11 & 3 & 0 \\
		LSTM D & ${\bf40}$ & 21 & 12 & 3 & 0 \\
		LSTM E & 36 & 30 & 7 & 0 & 0 \\
		\hline
		GT & 70 & 60 & 35 & 20 & 6 \\
		\hline\hline
		ZiPS & 93 & 59 & 18 & 8 & 0 \\
		\hline
		ZiPS GT & 260 & 114 & 60 & 43 & 8\\
		\hline
	\end{tabular}
	\caption{Number of correct predictions in 2019 with difference in 3.}
	\label{table:2019D3}
\end{table}

If we use a larger difference interval, Table \ref{table:2018D3} and \ref{table:2019D3} tell us that correct predictions enhance a lot for the classes under 29 home runs for all models. Increment for the classes is more than 30 home runs, however, is little. SVM in 2018 and 2019 can predict one player who hit more than 40 home runs. From these two tables, we felt surprised that it is so hard for every model to predict the correct results for players who can hit more than 20 HRs even in this bigger difference. For the problem, we think there should be more investigations. On the other hand, ZiPS has significant improvement in this difference which shows it can not predict exact players' performance but it has a reasonable estimation about that.

\begin{table}[h!]
	\centering
	\begin{tabular}[t]{|cccccc|}
		\hline
		& $\interval{0}{9}$ & $\interval{10}{19}$ & $\interval{20}{29}$ & $\interval{30}{39}$ & 40+ \\
		\hline
		Regression & 7 & 2 & ${\bf4}$ & 3 & 2 \\
		SVM & 19 & 14 & ${\bf4}$ & ${\bf2}$ & ${\bf0}$ \\
		RF & 7 & 9 & 22 & 7 & 2 \\
		NN & ${\bf0}$ & 55 & 34 & 11 & 2 \\
		\hline
		LSTM A & 5 & ${\bf0}$ & 14 & 6 & 2 \\
		LSTM B & 6 & 2 & 14 & 10 & 2 \\
		LSTM C & 4 & 2 & 7 & 3 & 2 \\
		LSTM D & 4 & 2 & 7 & 3 & 2 \\
		LSTM E & 7 & 2 & 5 & 7 & 2 \\
		\hline
		GT & 75 & 62 & 34 & 11 & 2 \\
		\hline\hline
		ZiPS & 51 & 15 & 5 & 10 & 1 \\
		\hline
		ZiPS GT & 273 & 97 & 56 & 21 & 2\\
		\hline
	\end{tabular}
	\caption{Number of correct predictions in 2018 with difference more than 10.}
	\label{table:2018DNA}
\end{table}

\begin{table}[h!]
	\centering
	\begin{tabular}[t]{|cccccc|}
		\hline
		& $\interval{0}{9}$ & $\interval{10}{19}$ & $\interval{20}{29}$ & $\interval{30}{39}$ & 40+ \\
		\hline
		Regression & 8 & 4 & 8 & 13 & 5 \\
		SVM & 17 & 19 & ${\bf7}$ & ${\bf5}$ & ${\bf3}$ \\
		RF & 3 & 19 & 28 & 15 & 4 \\
		NN & ${\bf0}$ & 55 & 35 & 20 & 6 \\
		\hline
		LSTM A & 16 & 10 & 9 & 8 & ${\bf3}$ \\
		LSTM B & 13 & 5 & 9 & 12 & 5 \\
		LSTM C & 14 & 5 & 8 & 8 & 4 \\
		LSTM D & 13 & 9 & 12 & 12 & ${\bf3}$ \\
		LSTM E & 9 & ${\bf2}$ & 10 & 12 & 5 \\
		\hline
		GT & 70 & 60 & 35 & 20 & 6 \\
		\hline\hline
		ZiPS & 62 & 7 & 12 & 20 & 6 \\
		\hline
		ZiPS GT & 260 & 114 & 60 & 43 & 8\\
		\hline
	\end{tabular}
	\caption{Number of correct predictions in 2019 with difference more than 10.}
	\label{table:2019DNA}
\end{table}

Finally, we would like to know how many differences between true answers and predictions are more than 10. Table \ref{table:2018DNA} and Table \ref{table:2019DNA} show that most methods can not predict players with more than 30 home runs well. Although performing poorly for home run less than 20, SVM can figure out players with 20 or more home runs well. LSTM does a good job when the home run is less, but it is a little weak for 30 or more home runs. SVM's performance may be considered to adjust LSTM models to make better prediction results. 

From all tables above, we can see that LSTM models have excellent ability to predict most players who hit less than 30 homers, but for classes more than 30, they may need more training data and information to enhance their ability. SVM does not make good jobs in the first two classes, but has a great ability to make correct predictions on the classes LSTM models do not perform well. The neural network is not a good way for this task because it focuses too much on the players with less than 19 home runs. Linear regression and ZiPS can not predict exactly true numbers, but they can include most of the difference within 10. Overall, LSTM is a good way to predict the true number and is better than machine learning ways and existing projection systems. 

\subsection{Analysis of players with 40 home runs}

\begin{table}[h!]
	\centering
	\begin{tabular}[t]{|ccccccc|}
		\hline
		& GT & Regr & SVM & RF & NN & ZiPS \\
		\hline
		Khris Davis & 48 & 30 & 43 & 28 & 0 & 36\\
		J.D. Martinez & 43 & 29 & 40 & 24 & 0 & $\times$ \\
		\hline\hline
		Nolan Arenado & 41& 33 & 33 & 33 & 0 & 36\\
		Nelson Cruz & 41 & 28 & 39 & 16 & 0 & 30\\
		Jorge Soler & 48 & 10 & 7 & 2 & 0 & 13\\
		Eugenio Suarez & 49 & 24 & 34 & 15 & 0 & 30\\
		Mike Trout & 45 & 33 & 39 & 39 & 0 & 38\\
		Christian Yelich & 44 & 24 & 29 & 13 & 0 & 28\\
		\hline
	\end{tabular}
	\caption{Predicitons for the players with 40 and more home runs.}
	\label{table:ML40}
\end{table}

There is another question we care about: how does each model's performance on players who can hit more than 40 home runs? If a model can figure out this class of player, it would provide additional benefits. From Table \ref{table:ML40} and \ref{table:LSTM40}, we list those players in 2018 and 2019 who can hit more than 40 home runs and their predictions by each model. We could find that SVM and ZiPS have closer results than other methods. LSTM models seem to be very conservative on their future performance. We show how many predictions are underestimated and overestimated in Table \ref{table:EST2018} and \ref{table:EST2019}. In both tables, we see that SVM and ZiPS have the most overestimated predictions, while LSTM models, random forest, and neural networks usually underestimate future performance. Therefore, we think that in the class of 40 or more home runs, predictions from SVM and ZiPS might tell us the maximum power of players, and random forest and LSTM would provide a lower bound of their performance.

\begin{table}[h!]
	\centering
	\begin{tabular}[t]{|ccccccc|}
		\hline
		& GT & A & B & C & D & E \\
		\hline
		Khris Davis & 48 & 23 & 21 & 27 & 26 & 26\\
		J.D. Martinez & 43 & 19 & 20 & 27 & 24 & 25\\
		\hline\hline
		Nolan Arenado & 41& 35 & 33 & 31 & 35 & 36\\
		Nelson Cruz & 41& 31 & 27 & 31 & 35 & 28\\
		Jorge Soler & 48& 21 & 14 & 17 & 9 & 11\\
		Eugenio Suarez & 49 & 24 & 25 & 25 & 24 & 21\\
		Mike Trout & 45& 35 & 33 & 31 & 35 & 32\\
		Christian Yelich & 44& 24 & 21 & 22 & 24 & 22\\
		\hline
	\end{tabular}
	\caption{LSTM Predicitons for the players with 40 and more home runs.}
	\label{table:LSTM40}
\end{table}

\begin{table}[h!]
	\centering
	\begin{tabular}[t]{|cccc|}
		\hline
		& Overestimated & Exact & Underestimated \\
		\hline
		ZiPS & 343 & 14 & 92\\
		Regression & 101 & 8 & 75 \\
		SVM & 115 & 16 & 53  \\
		RF & 64 & 10 & 110 \\
		NN & 0 & 5 & 179 \\
		\hline
		LSTM A & 66 & 22 & 96\\
		LSTM B & 65 & 25 & 94 \\
		LSTM C & 67 & 20 & 97 \\
		LSTM D & 69 & 19 & 96 \\
		LSTM E & 86 & 23 & 75 \\
		\hline
	\end{tabular}
	\caption{Predicitons overview in 2018.}
	\label{table:EST2018}
\end{table}

\begin{table}[h!]
	\centering
	\begin{tabular}[t]{|cccc|}
		\hline
		& Overestimated & Exact & Underestimated \\
		\hline
		ZiPS & 292 & 25 & 168\\
		Regression & 71 & 6 & 114 \\
		SVM & 99 & 15 & 77  \\
		RF & 43 & 13 & 135 \\
		NN & 0 & 4 & 187 \\
		\hline
		LSTM A & 72 & 18 & 101\\
		LSTM B & 72 & 24 & 95 \\
		LSTM C & 79 & 19 & 93 \\
		LSTM D & 55 & 21 & 115 \\
		LSTM E & 66 & 22 & 103 \\
		\hline
	\end{tabular}
	\caption{Predicitons overview in 2019.}
	\label{table:EST2019}
\end{table}

Here we get a closer look at players, Table \ref{table:201840} and \ref{table:201940} show their past five years performance. Time at bat (AB) means their chance to face pitchers in a season, and HR shows their previous power and talent. In these players, most of the players have stable performance in the previous five years except for Jorge Soler and Christian Yelich. All models have similar positive predictions on Nolan Arenado. However, LSTM models and random forest make lower prediction numbers than SVM and ZiPS in other players. It is our future work to make LSTM models make more reasonable predictions instead of minimum numbers for those players who can create at least 40 home runs. On the other hand, Jorge Soler and Christian Yelich are the hardest players to make correct predictions. Jorge Soler has limited opportunities in the previous five years, so it is really hard to imagine he could hit more than 40 home runs in the sixth year after he got enough AB (589 AB in 2019). Christian Yelich has equal AB in the past five years, but his home run numbers vary from 7 to 36. It is hard to judge his great performance would continue or drop down to previous performance.

\begin{table}[h!]
	\centering
	\begin{tabular}[t]{|cccccc|}
		\hline
		HR & 2013 & 2014 & 2015 & 2016 & 2017  \\
		\hline
		Khris Davis & 11 & 22 & 27 & 42 & 43 \\
		J.D. Martinez & 7 & 23 & 38 & 22 & 45 \\
		\hline\hline
		AB & 2013 & 2014 & 2015 & 2016 & 2017  \\
		\hline
		Khris Davis & 136 & 501 & 392 & 555 & 566 \\
		J.D. Martinez & 296 & 441 & 596 & 460 & 432 \\
		\hline
	\end{tabular}
	\caption{Previous performance of players who hit more than 40 home runs in 2018.}
	\label{table:201840}
\end{table}

\begin{table}[h!]
	\centering
	\begin{tabular}[t]{|cccccc|}
		\hline
		HR & 2014 & 2015 & 2016 & 2017 & 2018 \\
		\hline
		Nolan Arenado & 18& 42 & 41 & 37 & 38 \\
		Nelson Cruz & 40 & 44 & 43 & 39 & 37 \\
		Jorge Soler & 5 & 10 & 12 & 2 & 9 \\
		Eugenio Suarez & 4 & 13 & 26 & 26 & 34 \\
		Mike Trout & 36 & 41 & 29 & 33 & 39 \\
		Christian Yelich & 9 & 7 & 21 & 18 & 36 \\
		\hline\hline
		AB & 2014 & 2015 & 2016 & 2017 & 2018 \\
		\hline
		Nolan Arenado & 432& 616 & 618 & 606 & 590 \\
		Nelson Cruz & 613& 590 & 589 & 556 & 519 \\
		Jorge Soler & 89 & 366 & 227 & 97 & 223 \\
		Eugenio Suarez & 244 & 372 & 565 & 534 & 527 \\
		Mike Trout & 602 & 575 & 549 & 402 & 471 \\
		Christian Yelich & 582 & 476 & 578 & 602 & 574 \\
		\hline
	\end{tabular}
	\caption{Previous performance of players who hit more than 40 home runs in 2019.}
	\label{table:201940}
\end{table}

\subsection{Potential Impacts and Lessons Learned}
For the potential impacts, we have examined the new projection method by deep learning and analyzed the results to provide more details from the systems in the paper. The knowledge could be helpful for the domain users since they could get more accurate predictions and could get larger benefits from the information. In the past, domain experts, such as team managers, scouts, and coaches, in the baseball industries would convert their observations and experience into useful knowledge and are widely believed. However, The decisions made by domain experts may depend on their instincts or biases \cite{Schumaker2010}. Moreover, data is overloaded and may come from multimedia now, so data mining would be performed to make sense of the data and further help the domain experts \cite{schumaker2010sports}. The historical data could be used to justify the decision. Also, The usage of data mining skills could be claimed that the decisions are free from biases. Hence, with our results, team managers, scouts, and coaches can have a better understanding of the performance of players from the actionable knowledge \cite{schumaker2010sports}. Moreover, they could expect players’ future growth, draft players more effectively, and sign smarter contracts with actionable knowledge. With the actionable knowledge, organizations or groups could stay competitive with their opponents and take advantage of the information \cite{Schumaker2010}. Managers and coaches could use the technologies to simulate players' performance and make the most optimistic strategies for the coming or future season.

For a broad context, in our works, we not only put emphasis on the LSTM structures and application but also stand in the domain-driven point of view. Cao (2007, 2010) has proposed domain-driven data mining to solve increasingly complex challenges in the real world \cite{cao2007domain}, \cite{cao2010domain}. The data mining should not just present the result and center by data itself, but it should be useful for domain users and benefit organizations. Also, the actionable knowledge provided by domain-driven data mining could fit users’ needs and close the gap between researchers and practitioners \cite{pinheiro2020bi}. That was the lesson we learned that it is more important to create actionable knowledge for the organization through data mining, so we do not only focus on the deep learning application but also consider its real usage in the baseball domain. To be more specific, our experiments could directly benefit the domain users since they can make decisions based on the more accurate numbers for the players. They could get more close predicting numbers from LSTM than the old ones predicted by the existing projection systems and machine learning algorithms. The actionable knowledge could be taken by users and get advantages from the result. For example, they are able to carry the information to decide the salary for players and evaluate their potential in the future. Therefore, we have successfully collaborated human decisions and mining systems and improved baseball knowledge actionability, which are two features Kumari (2011) identified as features of domain-driven data mining, in our research \cite{kumari2012data}.
In conclusion, we believe our study became the bridge to connect academic development and domain needs.

\subsection{Limitations and Challenges}
For data limitation, we assume that players have the same height and weight in their career since we are not able to access their exact body information every year. Once we get their latest body information, we would use it as their career weight and height. Moreover, the other 18 features are all the basic information that shows players’ performance on the field, and since there did not exist too many previous researches, we decided to include all the possibilities and did not eliminate any information of stats. Therefore, reducing useless features could be a direction of future researches. Furthermore, we did not include the advanced statistics such as hard-hit rate and batting average on balls in play (BABIP) because they are not easy to get and old players had no such information. Although we believe the information would be useful, we can only leave it for future research. On the other hand, we would like to try a variety of models on the topic, but due to time limitations and lack of previous researches, we can only focus on our LSTM models and other simple baselines. More models could be tested in the later studies, such as GRU, BiLSTM, and AT-LSTM we have used in the paper. Other novel sequential models would also be a great target to test. Besides, finding how the existing projection system works is the biggest challenge for us. Those results are usually protected and should be paid to get, and their formulas are usually hidden or described in a blurred manner.

\subsection{Future Works}
For future researches, there would be four directions that we are interested in and believe would bring positive influence on the topic. First, it is helpful to break the data limitation in this paper. To be more specific, researchers could consider the year-to-year weight and height of players and add more biological information into the database. That gonna help the model give more accurate predictions. Also, it would be great to combine the advanced statistics with the current database to tell the models about the information of our players. Second, trying more deep learning models and comparing the results with each other can help us get profound knowledge of the topic. For example, transformer models and CNN-based approaches are great ideas. Their result could be compared with ours to judge the performance. Third, since we only focus on HR, it is encouraged to have more studies on different stats. We guess that each item may need different kinds of models to predict and this thought could only be verified by lots of experiments. Even more, we can create representations of players' performance based on these experiments and ours in order to simplify the evaluation process and find hidden patterns of players. Last but not least, we observed that players who hit more than 30 HRs were hard to predict by any models, so we are curious about the reasons for that. It is a hard job and needed more examinations, but we still believe these players are predictable based on our findings. Understanding the reasons could help researchers have more knowledge on the issue and have a chance to solve it.

\section{Conclusion}

In conclusion, we explore the possibility of using a deep learning model, LSTM, to make home runs prediction for those players in MLB. It is a new research direction on using deep learning to predict baseball players' future performance based on their past statistics.  Our results have shown that deep learning could be a new and better model to solve the projection problem and become a new system to provide valuable future information. We also had a comprehension analysis of our result and created an insightful aspect of the problem. In addition, we dig deeper to discover interesting views of baseball players. Hence, we believe our works could be marked as a new baseline of the problem and help the domain users.

In this paper, we propose 5 LSTM models and compare their capacity with several machine learning models and a widely used projection system, ZiPS. Our results show that LSTM models have low MAE and RMSE in 2018 and 2019. Although linear regression shows similar low RMSE and MAE in 2018, LSTM models outperform all methods in 2019. LSTM models also make the most correct predictions in small difference intervals, while other methods have a huge difference between their predictions and true home runs. Overall, we conclude that LSTM is a useful method to create home run predictions for MLB players. It is a robust way and can predict more accurately than machine learning methods and existing projection systems.



\bibliography{scibib}

\bibliographystyle{Science}


\end{document}